\documentclass[11pt]{article}

\usepackage[final]{acl}

\usepackage{times}
\usepackage{latexsym}

\usepackage[T1]{fontenc}

\usepackage{xspace}
\newif\ifshowcomments

\makeatletter
\let\todo\@undefined
\makeatother

\ifshowcomments
    \usepackage[textsize=scriptsize,textwidth=2.4cm]{todonotes}
\else
    \usepackage[disable,textsize=scriptsize,textwidth=2.4cm]{todonotes}
\fi

\usepackage{lineno}
\usepackage{amsmath}
\usepackage{amssymb}
\usepackage{mathtools}
\usepackage{amsthm}
 \usepackage{adjustbox}

\usepackage{multirow}
\usepackage{makecell}
\usepackage{graphicx}
\usepackage{hyperref}
\usepackage{url}
\usepackage{booktabs, multirow, array}
\usepackage{xcolor}
\usepackage{colortbl}
\usepackage{graphicx}
\usepackage{tabularx}
\usepackage{multirow}
\usepackage{threeparttable}
\usepackage{array}
\usepackage{multirow}
\usepackage{makecell}
\usepackage{comment}
\usepackage{makecell}
\usepackage{placeins}
\usepackage{listings}
\usepackage{xcolor}
\usepackage{enumitem}
\usepackage{wrapfig}
\usepackage{listings}
\usepackage{amsthm}
\usepackage{amsmath}

\lstdefinestyle{pythonstyle}{
    language=Python,
    basicstyle=\ttfamily\footnotesize,
    breaklines=true,
    showstringspaces=false
}
\definecolor{darkblue}{rgb}{0, 0, 0.5}
\hypersetup{colorlinks=true, citecolor=darkblue, linkcolor=darkblue, urlcolor=darkblue}
\usepackage{microtype}

\usepackage{url}
\usepackage{graphicx}
\usepackage{booktabs}
\usepackage{afterpage}
\usepackage{caption}
\usepackage{subcaption}
\usepackage{multirow}

\usepackage{multirow}
\usepackage{listings}
\usepackage{xcolor}
\usepackage{float}
\usepackage{graphicx} 

\usepackage{makecell}

\usepackage[utf8]{inputenc}
\usepackage{todonotes}

\usepackage{microtype}

\usepackage{inconsolata}
\usepackage{newfloat}
\usepackage{comment}

\usepackage{newfloat}
\usepackage{listings}
\usepackage{amsmath} 
\usepackage{xcolor}
\usepackage{listings}
\usepackage{microtype}
\usepackage{dblfloatfix}
\usepackage{multirow}

\definecolor{specbg}{RGB}{247,248,250}
\definecolor{specframe}{RGB}{205,211,220}
\definecolor{specink}{RGB}{35,39,47}
\lstdefinestyle{specjson}{
    basicstyle=\ttfamily\scriptsize\color{specink},
    backgroundcolor=\color{specbg},
    frame=single,
    rulecolor=\color{specframe},
    framerule=0.6pt,
    framesep=4pt,
    showstringspaces=false,
    breaklines=true,
    breakatwhitespace=false,
    columns=fullflexible,
    keepspaces=true,
    tabsize=2,
    aboveskip=1pt,
    belowskip=0pt
}

\usepackage{graphicx}
\usepackage{array}
\usepackage{tabularx}

\title{SABRE: \underline{S}calable and \underline{A}utomated \underline{B}enchmarking
of VLMs under St\underline{re}ss}

\usepackage{tgheros}
\newcommand{\name}{\textsf{SABRE}\xspace}

\author{
  Zixuan Lan$^{*, \dagger}$\\
  University of Chicago\\
  {\tt\small zixuanlan@uchicago.edu}
  \And
  Luzhe Sun$^*$\\
  Toyota Technological Institute at Chicago\\
  {\tt\small luzhesun@ttic.edu}
  \AND
  Matthew R. Walter\\
  Toyota Technological Institute at Chicago\\
  {\tt\small mwalter@ttic.edu}
  \And
  Jiawei Zhou\\
  Stony Brook University\\
  {\tt\small jiawei.zhou.1@stonybrook.edu}
}

\begin{document}
\maketitle

\begingroup
\renewcommand{\thefootnote}{*}
\footnotetext{Equal contribution.}
\renewcommand{\thefootnote}{$\dagger$}
\footnotetext{Work done as a research intern at Stony Brook University.}
\endgroup

\begin{abstract}

Vision-language models (VLMs) are improving rapidly, but benchmark development
lags behind, making weaknesses hard to identify. Building stress tests is
costly: samples must satisfy controlled conditions, remain answerable, and
challenge current models. We present \name{}, a scalable, automated pipeline
that converts a Test Primer(a Markdown Task Design with Data Schema) into
structured specifications, generated or edited images, and question--answer
pairs. Automated filtering removes
candidates solved by a Filtering VLM, while human review verifies candidate
validity and supports annotation correction and localized image repair. We instantiate \name{}-Prior to test whether VLMs follow visual evidence instead
of relying on world priors---learned expectations about familiar objects and
scenes. Its 600 images and 1,000 questions span \textbf{Context} (unexpected
entities in familiar scenes), \textbf{Texture} (counterfactual materials),
\textbf{Attribute} (noncanonical component counts), and \textbf{Language
Elicitation} (answers suggested by language but unsupported by the image).
Across six VLMs, macro-average accuracy ranges from 17.8\% to 31.3\%
(22.6\% mean). A real-image Attribute control is comparably difficult for the
Filtering VLM. \name{}-Counting and \name{}-Spatial pilots show that the
workflow supports other stress-test settings. These results establish \name{} as a reusable framework
for constructing and refreshing VLM stress tests rather than a single fixed
benchmark. Project website and resources will be available at \url{https://zesearch.github.io/vlm-SABRE/}.

\end{abstract}

\section{Introduction}

\begin{figure*}[t!]
    \centering
    \includegraphics[width=0.95\textwidth]{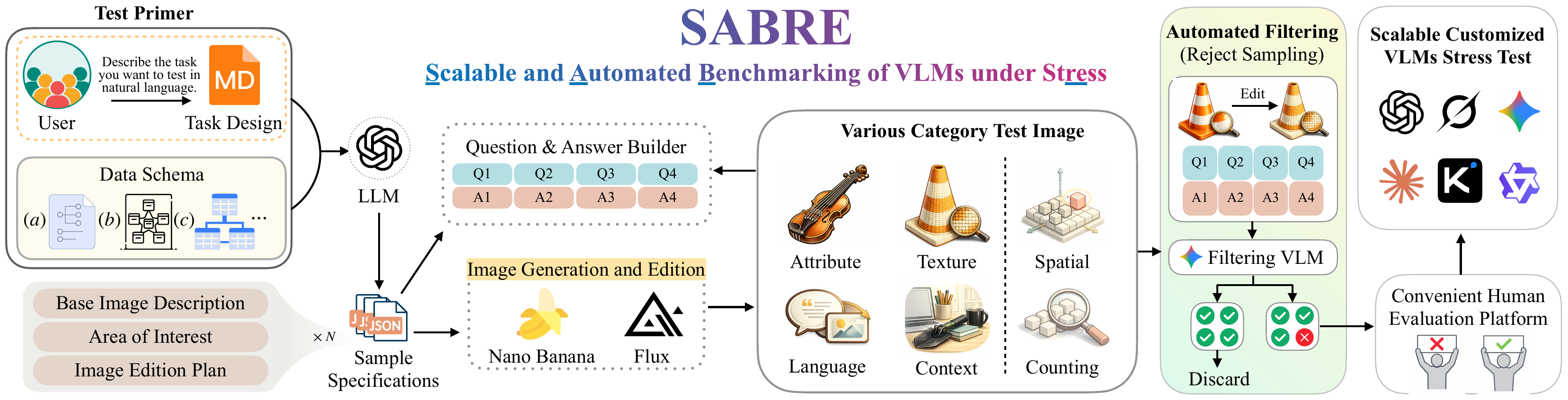}
    \caption{Overview of the \name{} benchmark-construction pipeline.}
    \label{fig:pipeline}
\end{figure*}

Vision-language models (VLMs) are advancing rapidly, while reliable benchmarks remain slow to design, construct, and validate\citep{fu2024mmesurveycomprehensivesurveyevaluation,ICLR2025_36d9468e}. As fixed benchmarks saturate, they reveal less about the weaknesses that remain and provide less guidance for further model development.
This paper asks a direct question: \textit{Can we construct targeted and reliable VLM stress tests fast enough to keep pace with model development?}

Large-scale visual benchmarks have traditionally been built by collecting real images and annotating them with human labor. ImageNet~\citep{5206848} and MS COCO~\citep{lin2015microsoftcococommonobjects} are landmark examples of this approach. This approach is especially costly for stress tests, which evaluates a target capability under deliberately demanding, distracting, or counterintuitive conditions while keeping the task valid and answerable.
\textbf{Building stress tests from real-image collections is slow and labor-intensive,} benchmark designers must devise enough rare and counterintuitive examples to probe the target failure mode systematically which makes it extremely hard to scale up.
To reduce annotation costs, several approaches automatically derive questions or labels from existing image collections. For example, POPE constructs object-existence questions from images and object annotations, while AutoConverter automatically turns existing visual questions into challenging multiple-choice questions \citep{li-etal-2023-evaluating,Zhang_2025_CVPR}. 
These approaches show that much of the annotation process can be automated. However, they still inherit the images from source datasets where the visual evidence in these images may not be sufficient for complicated stress test setting. Image generation alone does not produce a valid benchmark: generators may omit requested objects, alter unrelated regions, or produce incorrect counts\citep{10377168,huang2025t2icompbenchenhancedcomprehensivebenchmark, ghosh2023genevalobjectfocusedframeworkevaluating}, causing questions and reference answers derived from the benchmark specification to disagree with the actual image. Benchmark construction therefore requires both specification-faithful generation and explicit verification.

To address these challenges, we introduce a modular pipeline in which a benchmark designer provides a Test Primer which consists of a natural-language Markdown Task Design, a Data Schema that defines task-specific fields and validation rules, and question format, such as open generation or multiple choice. The pipeline converts this specification into structured sample specifications. It then generates the required images, applies task-oriented edits when needed, and builds the corresponding question--answer pairs. Each candidate is evaluated by a filtering VLM through a model-in-the-loop Automated filtering. Candidates solved by the filtering model are discarded, while those exposing a potential failure are retained. Because automatic filtering establishes difficulty but not validity, every retained candidate must undergo human verification.For tasks that require a specified image edit, reviewers compare the base image \(I^b\) with the edited image \(I^e\) to verify that the requested change is present and unrelated content is preserved. Reviewers can revise questions, reference answers, and region annotations, or repair localized image defects without regenerating the entire image. The same interface also accepts uploaded real images, which pass through the same screening and curation stages.

We first instantiate the pipeline as \name{}-Prior, which tests whether VLMs follow visual evidence when it conflicts with world priors, namely expectations learned from training data about how familiar objects, scenes, and situations normally appear\citep{liu2025phdchatgptpromptedvisualhallucination,
lee-etal-2025-vlind,pmlr-v267-luo25b,
vo2026visionlanguagemodelsbiased,
guan2024hallusionbenchadvanceddiagnosticsuite}. Such conflicts are underrepresented in conventional image collections because they are intentionally unusual and often require precise image edits. 

Nevertheless, these atypical scenes, edited content, and noncanonical object configurations can test a basic requirement of visual grounding: a model should answer from the image supplied by the user rather than from what it expects to see.
Using the pipeline, we construct \name{}-Prior across four complementary subsets: unexpected objects in familiar scenes (\textbf{Context}), answers suggested by question wording but unsupported by the image (\textbf{Language Elicitation}), counterfactual object materials (\textbf{Texture}), and noncanonical component counts (\textbf{Attribute}). To examine the pipeline in different stress-test settings, we also apply it to
\textbf{Counting} and \textbf{Spatial} reasoning using separate Test Primers. These two small-scale pilots are not intended as exhaustive evaluations on model capability; instead, they examine whether the pipeline can construct new types of stress tests without task-specific redesign.

We evaluate \name{}-Prior on six recent frontier VLMs. Across its four world-prior subsets, macro accuracy ranges from 17.8\% to 31.3\%, with an average of 22.6\% across models. The distinct performance profiles across subsets reveal complementary failures in following visual evidence and rejecting unsupported prior expectations.
Together, the \name{}-Prior results and the two pilot instantiations support the central contribution of this work: \name{} is not a single fixed benchmark, but a reusable framework for constructing, verifying, and refreshing targeted VLM stress tests as models evolve.

\section{Related Work}

\paragraph{VLM Evaluation and Stress Testing.}
VLM evaluation has expanded from task-specific VQA to broad suites covering
perception, reasoning, instruction following, and hallucination, such as MME,
MMBench, and MM-Vet, alongside targeted stress tests such as POPE and MMVP
\citep{fu2025mmecomprehensiveevaluationbenchmark,
liu2024mmbenchmultimodalmodelallaround,pmlr-v235-yu24o,
li-etal-2023-evaluating,Tong_2024_CVPR}.
These benchmarks show that strong aggregate performance can coexist with
systematic failures on carefully selected visual cases. However, most are fixed
evaluation sets focused on what to measure, whereas our work studies how
targeted stress tests can be constructed, screened, and maintained as models
and evaluation needs evolve.

\paragraph{Automated and Generative Benchmark Construction.}
Prior work reduces benchmark construction costs by automatically generating
questions, answers, and distractors from existing image collections. For
example, POPE derives object-existence questions from image annotations, while
AutoConverter transforms existing VQA questions into challenging
multiple-choice evaluations
\citep{li-etal-2023-evaluating,Zhang_2025_CVPR}. Although these methods automate
substantial annotation work, they cannot control the visual evidence contained
in the source images. Image generation, editing, and procedural rendering
address this limitation by directly constructing evaluation images.
ImageNet-D and JourneyBench use generated images to introduce controlled or
unusual visual conditions, while Vision-Language Bootstrapping, Auto-Comp, and
InfiniBench use dynamic image editing, synthetic image pairs, and customizable
3D scenes
\citep{Zhang_2024_CVPR,NEURIPS2024_734abb86,ICLR2025_36d9468e,
sbrolli2026autocompautomatedpipelinescalable,Wang_2026_CVPR}.
However, generative construction makes sample validity dependent on whether
each image realizes its intended specification. Our pipeline therefore
integrates structured sample specifications, image generation and targeted
editing, pressure screening, human verification, annotation revision, and
localized repair. It also supports generated images and uploaded real images
within the same curation workflow.

\paragraph{World Priors and Visual Evidence Conflicts.}
A growing body of work studies whether VLMs follow visual evidence when it conflicts with language or world knowledge. The PhD benchmark includes a counter-common-sense subset, PhD-ccs, that uses AI-generated images violating everyday expectations to evaluate visual hallucination \citep{liu2025phdchatgptpromptedvisualhallucination}. VLind-Bench measures language priors using DALL-E~3-generated counterfactual images together with auxiliary tests of commonsense knowledge, visual perception, and commonsense bias, thereby reducing alternative explanations for a model's failure \citep{lee-etal-2025-vlind}. ViLP constructs out-of-distribution image--question--answer triplets in which one condition can be answered from textual priors while the others require evidence from generated images \citep{pmlr-v267-luo25b}. VLMBias makes controlled counterfactual modifications to familiar subjects, such as well-known logos, animals, flags, and game pieces, to test whether memorized canonical knowledge overrides visual counting and identification \citep{vo2026visionlanguagemodelsbiased}. HallusionBench is broader than a world-prior benchmark: it uses human-collected and human-edited control pairs to diagnose both language hallucination and visual illusion, although some edited cases explicitly place visual content in conflict with common knowledge \citep{guan2024hallusionbenchadvanceddiagnosticsuite}. These benchmarks directly motivate our world-prior case study, and we include them as comparison benchmarks in our experiments. Our primary goal, however, is not to introduce another isolated test of prior reliance, but to use this demanding setting to instantiate and evaluate a general pipeline for scalable stress-test construction.

\section{Method}
\label{sec:method}

\begin{figure*}[t!]
    \centering
    \includegraphics[width=\textwidth]{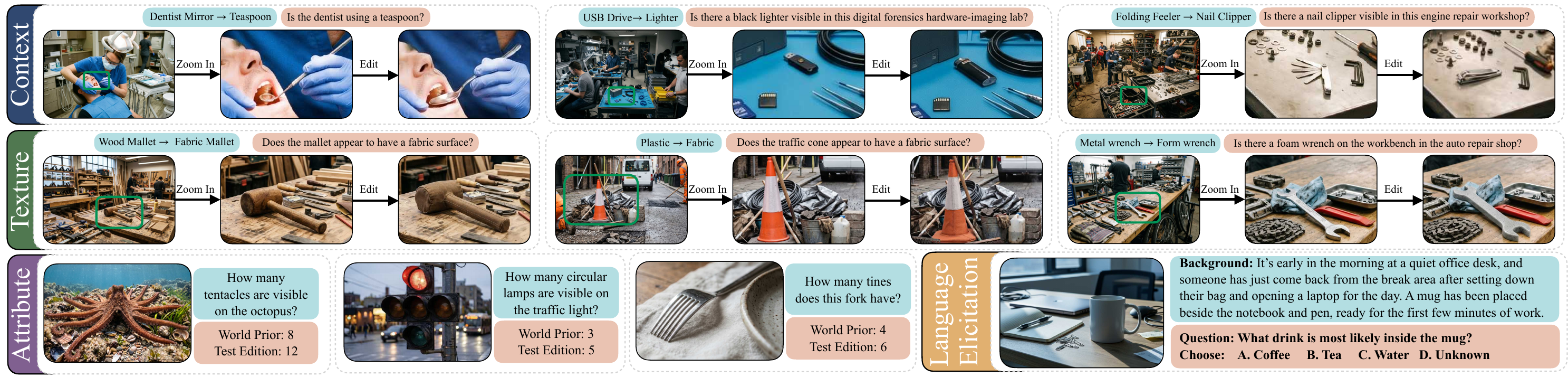}
    \caption{Samples of \name{}-Prior in 4 different subsets.}
    \label{fig:sample}
\end{figure*}

\subsection{Overview and Problem Formulation}
\label{sec:method_overview}

Our goal is to automatically create a VLM stress-test benchmark based on the pipeline. The pipeline takes a \texttt{Test Primer}, denoted by \(S\), which consists of a natural-language Markdown \texttt{Task Design}, a \texttt{Data Schema} that defines task-specific fields and validation rules, and question format, such as open generation or multiple choice. Given \(S\), the pipeline automatically constructs a set of \texttt{Candidate Samples} before human verification:
\[
\mathcal{C}(S)=\{x_i\}_{i=1}^{N}.
\]
Each candidate \(x_i=(I_i,q_i,y_i)\) consists of an image \(I_i\), a question \(q_i\), and its reference answer \(y_i\).

As illustrated in Figure~\ref{fig:pipeline}, an LLM first converts the test primer \(S\) into \texttt{Sample Specifications}. Each sample specification is a JSON file containing the information required by the subsequent steps, such as image generation and editing prompts. Let \(z_i\) denote the \(i\)-th sample specification. An image generation model, denoted by \(G\), generates the required images. When a design requires an image modification, an image editing model, denoted by \(E\), edits the generated image. After candidate set construction, the candidate samples enter \texttt{Automated Filtering}. A VLM, denoted by \(M\) and called the \texttt{Filtering VLM}, serves as the \texttt{Automated Filtering} and evaluates each candidate sample. Candidate samples answered correctly by \(M\) are filtered out, while those answered incorrectly are retained and sent to our Convenient Human Evaluation Platform, for human verification and image repair. Candidate Samples that pass human verification become \texttt{Samples} and are included in the final benchmark.

\subsection{Modular Task Recipes}
\label{sec:task_specification}

A test primer \(S\) defines one stress-test topic. It contains a natural-language Markdown task design, an data schema, and a selected question format. The Markdown file states what to test and how to construct the images. It can specify the scene, the visual content to add, remove, or modify, and the content that must remain unchanged. The question format specifies how the VLM should answer, such as open generation or multiple choice.

The data schema controls the structure of the JSON output. It defines the data types (e.g., text or lists) and other constraints. The pipeline uses the default data schema when the task recipe does not provide a custom one.

A design LLM reads \(S\) and produces sample specifications in JSON. Each sample specification contains the information needed to construct one candidate sample, including an image generation or editing prompt, a question, and its reference answer. 

A new stress-test topic only requires a new test primer. The pipeline reuses the same image generation, pressure screening, human verification, image repair, and export workflow for all topics. This separation provides extensibility without requiring a new benchmark-construction pipeline for each topic.

\subsection{Automatic Candidate Sample Construction}
\label{sec:candidate_construction}

Each sample specification \(z_i\) contains an image prompt, a question \(q_i\), and a reference answer \(y_i\). For a single-image task, the image generation model \(G\) generates an image \(I_i\) from the prompt \(p_i\):
\[
I_i=G(p_i).
\]

Some stress-test topics test whether a VLM responds to a specific visual change, such as removing an object or changing an object attribute. These topics require image editing. The edit changes the target visual evidence while keeping the rest of the scene unchanged. This isolates the target change and tests whether the VLM updates its answer based on that change.

For an editing task, \(z_i\) contains a base-generation prompt \(p_i^b\) and an editing prompt \(p_i^e\). The pipeline uses \(G\) to generate the base image \(I_i^b\) and the image editing model \(E\) to produce the edited image \(I_i^e\):
\[
I_i^b=G(p_i^b),
\qquad
I_i^e=E(I_i^b,p_i^e).
\]
The editing prompt states what to change and what to preserve. A real image can replace \(I_i^b\), which skips the image generation step.

The pipeline pairs each applicable image with one question and its reference
answer. Each sample specification produces exactly one candidate sample:
\[
x=(I,q,y).
\]
After human verification, each accepted sample forms or joins a case. A case
contains one sample for one-to-one tasks and multiple samples when several
image--question pairs derive from the same underlying image construction or
visual intervention. The pipeline stores the prompts, model identifiers, and
generation records as metadata.

\subsection{Model-in-the-Loop Pressure Screening}
\label{sec:pressure_screening}

Automated filtering removes candidate samples that the Filtering VLM \(M\) answers correctly and retains those that it answers incorrectly. For each candidate sample \(x_i=(I_i,q_i,y_i)\), \(M\) produces a response
\[
r_i=M(I_i,q_i).
\]
The evaluator compares \(r_i\) with the reference answer \(y_i\) using the scoring rule for the selected question format:
\[
e_i=\operatorname{Eval}(r_i,y_i),
\qquad e_i\in\{0,1\},
\]
where \(e_i=1\) indicates a correct response and \(e_i=0\) indicates an incorrect response.

The automated filtering retains only the candidate samples that \(M\) answers incorrectly:
\[
\mathcal{C}_{\mathrm{pressure}}(S;M)
=
\{x_i\in\mathcal{C}(S)\mid e_i=0\}.
\]

The filtering VLM measures candidate sample's difficulty, not candidate sample's validity. An incorrect response from \(M\) may reveal a genuine model failure, but it may also result from an unsuccessful image generation or edit, an ambiguous question, or an incorrect reference answer. Every retained candidate sample must therefore undergo human verification before it can become a sample.

\subsection{Human Verification and Localized Image Repair}
\label{sec:data_curation}

Candidate samples retained by the automated filtering enter human verification platform. Reviewers inspect the image \(I_i\), question \(q_i\), reference answer \(y_i\), sample specification \(z_i\). A candidate sample passes verification only when the required visual evidence is present, the requested edit is correctly applied when needed, the question is unambiguous, and the reference answer matches the image. Reviewers can accept or reject a candidate sample, revise its question or reference answer, or repair a local image defect.

For image repair, the reviewer selects a bounding box \(b_i\) and provides a repair instruction \(u_i\). The platform expands the selected region to include its surrounding visual context and crops the resulting patch \(P_i\) from the image. It marks the target region inside the patch and sends the marked patch, rather than the full image, to Gemini 3.1 Flash Image Preview with \(u_i\). Restricting the model input to a local patch focuses the edit on the selected defect and prevents unintended changes to the rest of the image.

The repaired patch \(\widetilde{P}_i\) cannot be directly pasted into the image because a hard boundary may introduce visible color and texture discontinuities. We therefore construct a soft mask \(A_i\) by expanding the target region and applying Gaussian blur to its boundary. The platform blends the repaired and original patches as
\[
P_i^\star
=
A_i\odot\widetilde{P}_i
+
(1-A_i)\odot P_i.
\]
It then places \(P_i^\star\) back into the original image. The center of the mask preserves the repaired content, while the smooth boundary gradually transitions to the original pixels. This soft blending reduces rectangular seams and integrates the repaired region more naturally into the surrounding scene.

\begin{figure}
    \centering
    \includegraphics[width=1.0\linewidth]{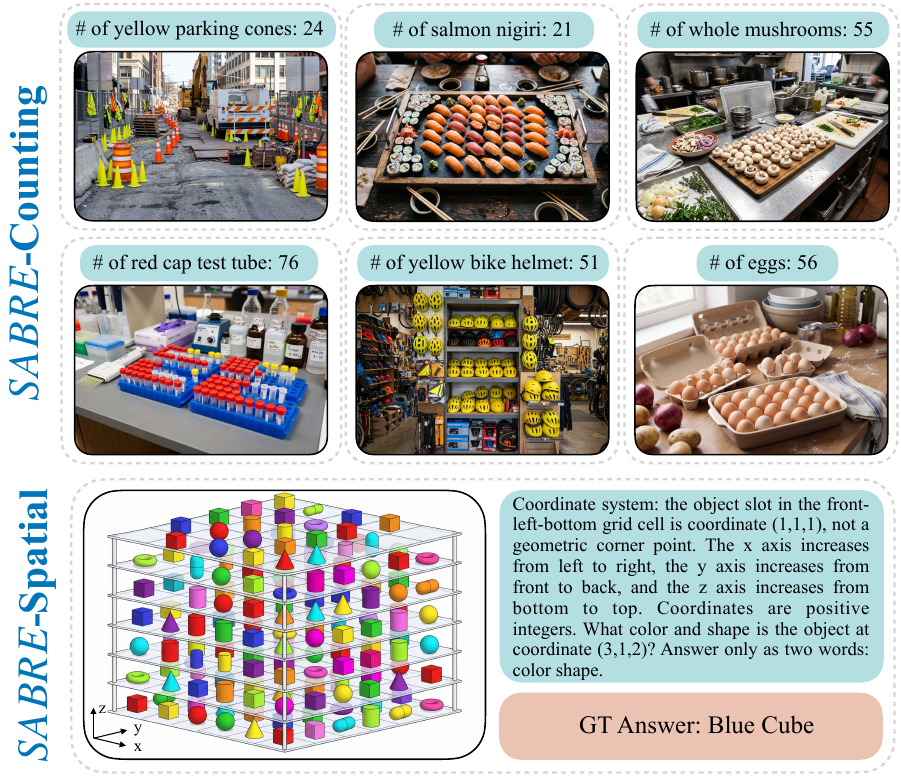}
    \caption{\name{}-counting and \name{}-spatial}
    \label{fig:pilot}
\end{figure}

The reviewer compares the repaired image with the original and may accept it, repair it again, or reject the candidate sample. Our questionnaire study shows that the localized repair tool produces more successful and visually natural repairs than direct whole-image Gemini editing and the other evaluated methods. Participants consistently preferred the results produced by our tool. The full comparison and questionnaire results are provided in the appendix.

Let \(x_i^\star\) denote the candidate sample after human verification and any reviewer revisions or image repairs. The reviewer assigns a validity decision \(v_i\in\{0,1\}\), where \(v_i=1\) indicates acceptance. The final benchmark is
\[
\mathcal{D}(S;M)
=
\left\{
x_i^\star
\mid
x_i\in\mathcal{C}_{\mathrm{pressure}}(S;M),
\ v_i=1
\right\}.
\]
Each item in \(\mathcal{D}(S;M)\) originates from a pressure-selected candidate sample and passes human verification. We call each retained item a sample.

The curation platform also supports real-image authoring. Users can upload a real image, provide or revise its question and reference answer, and apply localized image repair when needed. 
\section{Benchmark Instantiation}
\label{sec:world_prior}

\begin{figure*}[t!]
    \centering
    \includegraphics[width=\textwidth]{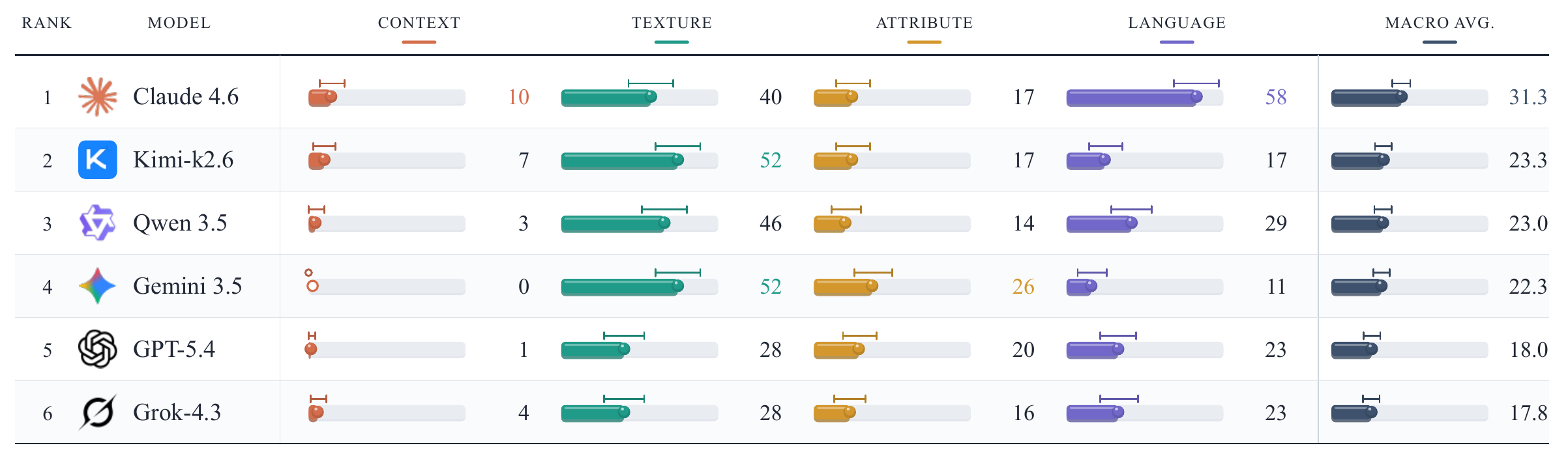}
    \caption{
\textbf{Accuracy (\%) of six frontier VLMs on \name{}-Prior.}
Bars show point estimates, and whiskers indicate the corresponding confidence
intervals. Context and Texture report strict All4 accuracy: a case is
correct only when all four Base--Edited yes/no probes are answered correctly.
Attribute uses normalized exact-match counting accuracy, while
Language Elicitation uses four-option multiple-choice accuracy. Macro Average is
computed across the four subset accuracies.
}
    \label{fig:main}
\end{figure*}

We first use our pipeline to construct \name{}-Prior, a benchmark that tests whether VLMs follow visual evidence or learned expectations about what is usually true. We refer to these learned expectations as world priors.

\paragraph{\name{}-Prior Stress-Test Subsets.}
The benchmark contains four complementary subsets. \textbf{Context} places unexpected objects in familiar scenes, such as replacing a microscope in a laboratory with a toaster. \textbf{Language Elicitation} uses question wording to suggest an answer that is unsupported by the image, such as asking what is inside a closed box. \textbf{Texture} assigns familiar objects counterfactual materials, such as giving a ceramic mug a furry surface. \textbf{Attribute} modifies canonical component counts while preserving object identity, such as showing a chair with five legs.

Context and Texture use paired base--edit images with four complementary yes/no probes. Language Elicitation uses single-image, four-option questions whose correct answer is \emph{unknown}, while Attribute uses open-ended counting questions over edited images.

\paragraph{\name{}-Prior Construction, Scale, and Evaluation.}

All four subsets use the shared workflow in Section~\ref{sec:method}. For each subset, GPT-5.4 reads its task primer \(S\) and produces sample specifications. FLUX.2 [flex], FLUX.2 [pro] \citep{blackforestlabs2025flux2}, and Gemini 3.1 Flash Image (Nano Banana 2) \citep{raisinghani2026nanobanana2} generate or edit the images specified by these specifications. Gemini 3.5 Flash serves as the filtering VLM \(M\). Candidate samples retained by the automated filtering undergo human verification and localized repair when needed. Each subset contains 100 cases. The final benchmark contains 400 cases, 600 images, and 1,000 samples, with one question per sample.

Context and Texture are evaluated using paired Base and Edited images. \textbf{Context} asks whether the source and target objects are visible in each image, with
expected answers \emph{yes/no} for Base and \emph{no/yes} for Edited. \textbf{Texture}
uses the same four-probe structure for the expected and counterfactual
materials, again producing \emph{yes/no} for Base and \emph{no/yes} for
Edited. A case is counted as correct only when all four yes/no samples are
correct. \textbf{Attribute} presents a single edited image with an open-ended
component-counting question; the extracted numeric response must exactly match
the annotated visible count after normalizing digits and number words.
\textbf{Language Elicitation} uses a single-image, four-option question whose wording
suggests an answer that cannot be verified visually; the correct option is
\emph{unknown}, with its position balanced across the dataset. We report each
subset's primary accuracy and their macro-average as the overall score.

\begin{figure*}[t!]
    \centering
    \includegraphics[width=0.95\textwidth]{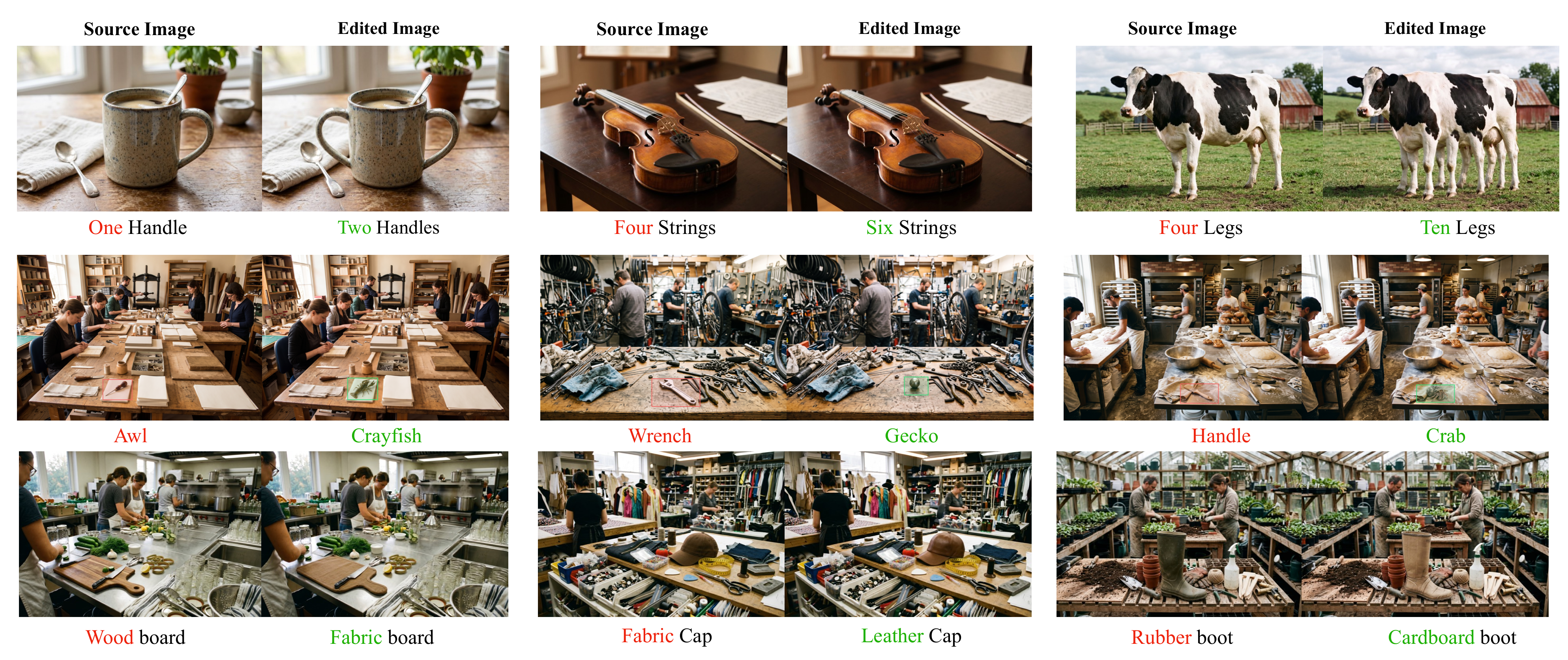}
    \caption{Examples of generated images from \name{}}
    \label{fig:sample1}
\end{figure*}

\begin{figure*}[t!]
    \centering
    \includegraphics[width=0.95\textwidth]{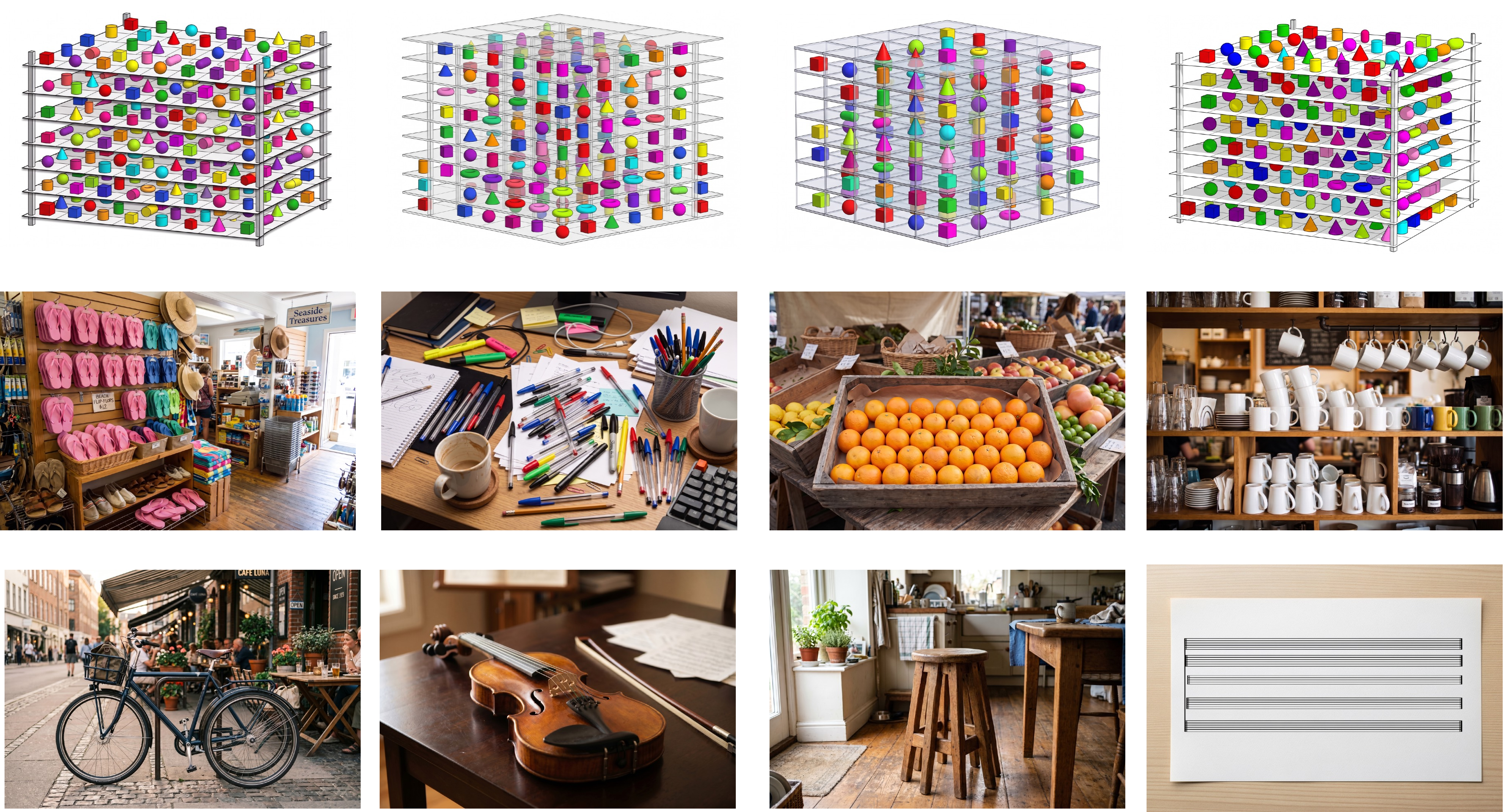}
    \caption{Additional examples of generated images from \name{}.
The first row shows Spatial QA instances, where the question-answer format
follows the same structure as illustrated in Figure~\ref{fig:pilot}.
The second row presents Counting instances, including examples that require
counting pink slippers, pens, oranges, and mugs from left to right.
The third row provides additional Attribute instances.}
    \label{fig:sample2}
\end{figure*}

\subsection{Additional Pipeline Instantiations}
To evaluate the extensibility of our pipeline, we use the same workflow to construct two additional 20-example pilots: Counting and Spatial. Counting presents dense scenes containing dozens of target objects and visually similar distractors, and asks the model to report the
number of visible target objects. Responses are scored by normalized exact count. Spatial presents dense multi-layer 3D lattices and includes multiple question formats, such as identifying the color and shape at a specified coordinate or counting objects that satisfy spatial and visual constraints, for example red cubes on the top layer. Responses are scored by normalized exact attribute matching or exact count, depending on the question type.

\section{Experiments}
\subsection{Experimental Setting}
\label{subsec:exp-setting}

% \begin{figure*}[t!]
%     \centering
%     \includegraphics[width=0.95\textwidth]{Figures/sample1.pdf}
%     \caption{Examples of generated images from \name{}}
%     \label{fig:sample1}
% \end{figure*}

% \begin{figure*}[t!]
%     \centering
%     \includegraphics[width=0.95\textwidth]{Figures/sample2.pdf}
%     \caption{Additional examples of generated images from \name{}.
% The first row shows Spatial QA instances, where the question-answer format
% follows the same structure as illustrated in Figure~\ref{fig:pilot}.
% The second row presents Counting instances, including examples that require
% counting pink slippers, pens, oranges, and mugs from left to right.
% The third row provides additional Attribute instances.}
%     \label{fig:sample2}
% \end{figure*}

\begin{table*}[t]
\centering
\small
\setlength{\tabcolsep}{6pt}
\renewcommand{\arraystretch}{1.08}
\begin{tabular*}{\textwidth}{@{\extracolsep{\fill}}lcllr}
\toprule
\textbf{Benchmark} & \textbf{Images} & \textbf{Construction Method} &
\textbf{Question Types} & \textbf{Gemini 3.5 Flash Acc. (\%)} \\
\midrule
PhD-CCS             & 750   & DALL-E 3                   & Yes/No                              & 82.3 \\
VLind-Bench         & 842 & DALL-E 3          & Yes/No                              & 90.0 \\
ViLP                & 900   & DALL-E 3 / FLUX            & Open identification                 & 70.3 \\
HallusionBench      & 181   & Manual editing             & Yes/No                              & 81.6 \\
VLMBias             & 1,392 & Semi-automated generation  & Counting and Yes/No                 & 60.3 \\
\midrule
\textbf{\name{}-Prior} & \textbf{600} & \textbf{Automated generation} &
\textbf{All above and MCQ} & \textbf{22.3} \\
\bottomrule
\end{tabular*}
\caption{
Same-model comparison using each benchmark's primary evaluation protocol
}
\label{tab:benchmark_difficulty}
\end{table*}

We evaluate six frontier VLMs on \name{}-Prior: Gemini 3.5 Flash \citep{googledeepmind2026gemini35flash}, GPT-5.4 \citep{openai2026gpt54}, Claude 4.6 Sonnet \citep{anthropic2026claudesonnet46}, Kimi-k2.6 \citep{moonshotai2026kimik26}, Qwen 3.5 27B \citep{qwen2026qwen35}, and Grok-4.3 \citep{xai2026grok43}. These models represent the current state of the art, and our stress tests are designed to identify their blindspots for future improvement. All models are evaluated with zero-shot prompts using greedy decoding (i.e., temperature $=0$) when supported. We estimate 95\% confidence intervals using a percentile bootstrap with 20,000 resamples, resampling cases within each subset. For each resample, we recompute every \name{}-Prior subset accuracy and their macro-average, defined as the unweighted mean of the subset accuracies. Detailed model identifiers, access dates, prompts, and response-parsing rules are provided in the Appendix.

\paragraph{Baselines} We additionally evaluate VCD \citep{10657718} and SoM \citep{yang2023setofmarkpromptingunleashesextraordinary} on the open-source Qwen 3.5 27B model to examine whether visual-enhancement methods improve performance. We also construct a real-image variant of Attribute subset to test whether model failures persist when the edited source images are real rather than generated. We also show more examples shown in \ref{fig:sample1} and \ref{fig:sample2}.

\subsection{Main Results}
\label{subsec:main-results}

Figure~\ref{fig:main} reports the primary subset scores of the six evaluated VLMs. Macro-average accuracy ranges from 17.8\% to 31.3\%, with a mean of 22.6\% across models. Because \name{}-Prior uses pressure screening and evaluates different behaviors with different question formats, the macro-average summarizes performance on this benchmark rather than providing a general ranking of model capability. Table~\ref{tab:benchmark_difficulty} provides complementary evidence. Gemini 3.5 Flash achieves 60.3\%--90.0\% accuracy on five existing benchmarks but only 22.3\% on \name{}-Prior. Its low score on \name{}-Prior partly reflects its role as the Filtering VLM. However, the other five models, which were not used during pressure screening, also score below 32\%. These results show that the constructed cases expose failures across frontier VLMs rather than only the Filtering VLM.

Context obtains the lowest scores under the strict  criterion: no
model exceeds 10\%, and the mean is 4.2\%. This reflects the difficulty of
consistently recognizing both the original and counterfactual objects across
the Base--Edit intervention. Texture scores are higher, ranging from 28\% to
52\%, suggesting that visible material evidence is more accessible than
context-incongruent object identity, although canonical material associations
still exert substantial pressure. Attribute ranges from 14\% to
26\%, combining fine-grained structural perception and exact counting with the
need to override canonical component counts. Language Elicitation shows the widest variation, from 11\% to 58\%. This subset
should not be interpreted as a pure measure of visual recognition: the queried
information is intentionally unavailable in the image. Instead, it tests
whether a model recognizes insufficient visual evidence and rejects a
plausible answer induced by the question wording. The large variation therefore
reveals differences in visual grounding, prior reliance, and willingness to
abstain.

Models perform very differently across subsets. Claude 4.6 achieves the highest macro-average, largely because of its strong Language Elicitation score. Gemini 3.5 Flash and Kimi-k2.6 perform best on Texture, while Gemini 3.5 Flash leads on Attribute but answers no Context cases correctly. The model ranking therefore changes across subsets, and no model performs best on all of them. These results show that \name{}-Prior exposes distinct model weaknesses that a single aggregate score would hide.

\subsection{Comparison with Existing Benchmarks}
\label{subsec:benchmark-comparison}

We further compare \name{}-Prior with five related benchmarks using Gemini 3.5 Flash. As shown in Table~\ref{tab:benchmark_difficulty}, the model achieves between 60.33\% and 90.02\% on PhD-CCS, VLind-Bench, ViLP, HallusionBench, and VLMBias, but only 22.30\% on \name{}-Prior. In particular, its accuracy exceeds 80\% on three of the existing benchmarks, suggesting that many previously constructed world-prior and hallucination cases are handled reliably by recent frontier models. In contrast, the cases produced by our pipeline continue to expose substantial failures.

The benchmarks use different question formats and scoring rules, so their accuracy values are not directly equivalent. However, the results show that several existing benchmarks provide limited headroom for a current frontier VLM. The low Gemini 3.5 Flash score on \name{}-Prior partly reflects its role as the Filtering VLM, but it also demonstrates that our pipeline can construct new cases that target weaknesses not exposed by increasingly saturated benchmarks. Unlike a fixed test set, our pipeline can generate new stress-test candidates on demand using updated task recipes and Filtering VLMs. This allows the evaluation to adapt as models improve. The pipeline supports flexible question formats, including yes/no, multiple-choice, and open-ended counting questions. It can therefore produce scalable stress tests across different task formats rather than remaining tied to one predefined benchmark.

\begin{table}[t]
\centering
\small
\renewcommand{\arraystretch}{1.05}
\begin{tabular*}{\columnwidth}
{@{\extracolsep{\fill}}lrrrrr@{}}
\toprule
\textbf{Method}
& \textbf{Ctx}
& \textbf{Tex}
& \textbf{Att}
& \textbf{Lang}
& \textbf{Avg.} \\
\midrule
Qwen 3.5 27B & 3 & 46 & 14 & 29 & 23.0 \\
+ VCD         & 1 & 34 & 16 & 27 & 19.5 \\
+ SoM         & 6 & 24 & 12 & 25 & 16.8 \\
\bottomrule
\end{tabular*}
\caption{\textbf{Effect of visual-enhancement methods on Qwen 3.5 27B.}
Results use the same four subset metrics as Figure~\ref{fig:main};
Avg. denotes their macro-average.}
\label{tab:visual_enhancement}
\vspace{-0.5em}
\end{table}

\subsection{Effect of Visual-Enhancement Methods}
\label{subsec:visual-enhancement}

A useful stress-test benchmark should not only expose model failures but also measure whether proposed methods fix them. We use \name{}-Prior to evaluate two visual-enhancement methods on Qwen 3.5 27B. Visual Contrastive Decoding (VCD) contrasts predictions from original and perturbed visual inputs, while Set-of-Mark (SoM) adds region markers to improve visual grounding. We evaluate both methods using the same questions and subset metrics as the standard Qwen 3.5 27B setting.

As shown in Table~\ref{tab:visual_enhancement}, neither method consistently improves performance. VCD and SoM improve selected subsets but reduce accuracy on others, resulting in lower macro-averages than the standard model. These results indicate that the failures exposed by \name{}-Prior cannot be fixed simply by highlighting visual regions or changing the decoding signal.

This experiment also demonstrates the diagnostic value of \name{}-Prior. It shows where a mitigation method helps, where it introduces regressions, and which model weaknesses remain unresolved. The benchmark can therefore support the development and evaluation of methods designed to address specific VLM failures.

\begin{table}[t]
\centering
\small
\setlength{\tabcolsep}{6pt}
\renewcommand{\arraystretch}{1.08}
\begin{tabular}{lcc}
\toprule
\textbf{Image Source} & \textbf{Correct / Total} & \textbf{Accuracy} \\
\midrule
Generated (Attribute  Modify) & 26 / 100 & 26\% \\
Real-Image Control        &  6 / 20  & 30\% \\
\bottomrule
\end{tabular}
\caption{\textbf{Real-image control.}
Exact-match counting accuracy of Gemini 3.5 Flash on generated and
real-image Attribute samples.}
\label{tab:real_image_control}
\end{table}

\subsection{Real-Image Control}
\label{subsec:real-image-control}

We test whether the low model accuracy on \name{}-Prior is caused by generated-image artifacts. We construct 20 control cases from real images using the same editing and evaluation procedure as the Attribute. As shown in Table~\ref{tab:real_image_control}, Gemini 3.5 Flash achieves 30\% accuracy on the real-image cases, compared with 26\% on the generated-image cases.

The similarly low accuracies suggest that the benchmark difficulty is not primarily caused by generated-image artifacts. They also support the reliability of the generated stress tests, which expose model failures at a level comparable to tests constructed from real photographs. The pipeline supports both real and generated source images, while generated images provide greater scalability and more precise control over the visual content being tested.

\begin{figure}[t]
    \centering
    \includegraphics[width=\columnwidth]{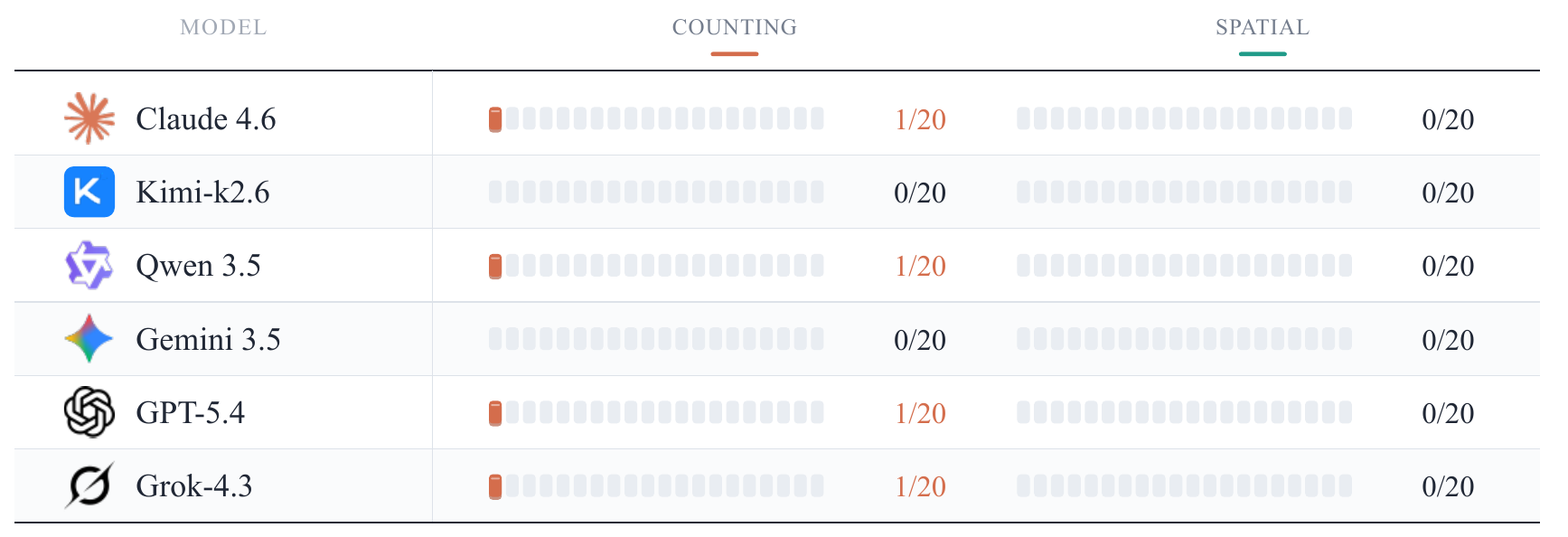}
    \caption{\textbf{Results on two additional stress-test instantiations.} Each cell represents one sample, and colored cells indicate correct model responses.}
    \label{fig:pilot_generalization}
\end{figure}

\subsection{Additional Stress-Test Instantiations}
\label{subsec:additional-instantiations}

We use two different test primers, to construct additional stress tests for counting and spatial reasoning. Each stress test contains 20 samples and uses the same generation, pressure screening, human verification, and repair workflow. Figure~\ref{fig:pilot_generalization} shows that both stress tests are challenging for all six VLMs. Each model answers at most one of the 20 Counting samples correctly, and none answers any Spatial sample correctly. These results contrast with the strong performance of frontier VLMs on many established benchmarks, which can become saturated as models improve. They also demonstrate the extensibility of our pipeline: different task specifications can be used to construct challenging stress tests for new visual capabilities without changing the shared workflow.

\section{Conclusion and Limitations}

We introduced \name{}, a scalable and modular pipeline for constructing VLM
stress tests from natural-language task specifications. The pipeline converts
each task specification into structured case specifications, constructs the
required images and question--answer pairs, retains challenging candidates
through failure-only pressure screening, and supports human verification and
localized repair. Using this pipeline, we constructed \name{}-Prior, covering four forms of conflict between visual evidence and world priors. Six recent VLMs achieve only 17.8\%--31.3\% macro accuracy and
exhibit distinct failure profiles. Representative Inference-time mitigation methods provide no consistent improvement, suggesting that the benchmark exposes persistent weaknesses in using visual evidence to override prior expectations rather than errors that can be removed by generic inference-time interventions. We also apply the same pipeline to Counting and Spatial
reasoning through separate task specifications, demonstrating its use across
different stress-test settings. A limitation shared by any fixed benchmark suite is that its coverage reflects
a finite set of task specifications at the time of release. The modular design
of \name{} mitigates this limitation by allowing new task specifications and
construction modules to be added as VLM capabilities evolve.

\bibliography{custom}

\newpage
\appendix
\onecolumn

\section{Detailed Experimental Setup}
\label{app:experimental-setup}

\paragraph{Evaluated Models.}
Table~\ref{tab:model_details} lists the exact model identifiers and inference
interfaces used in our experiments. All models were accessed on July 20, 2026.
API models were evaluated using the versions exposed under these identifiers
on that date. Each model received one image--question sample at a time without
in-context examples.

\paragraph{Inference Prompts.}
We use zero-shot, format-specific prompts and do not request explanations or
chain-of-thought reasoning. Context and Texture use the following prompt:
\begin{quote}
\small\ttfamily
Question: \{question\}\\
Answer with only one word: yes or no.
\end{quote}

Attribute and Counting use:
\begin{quote}
\small\ttfamily
Question: \{question\}\\
Answer with only the count or shortest possible count phrase. Do not explain.
\end{quote}

Language Elicitation presents the question followed by four options labeled
A--D and uses:
\begin{quote}
\small\ttfamily
Question: \{question and options\}\\
Answer with only the option letter. Do not explain.
\end{quote}
Spatial uses the same short-answer instruction as Counting. Depending on the
question, the expected response is either a count or a short combination of
attributes such as a color and shape. For the local Qwen evaluation, these
instructions are wrapped in the model's standard chat template with a
format-only system message requesting no reasoning or additional text.

\paragraph{Decoding and Image Input.}
The maximum output length is restricted to 8--16 tokens for API models,
depending on the response format. Qwen uses at most 32 newly generated tokens.
Images are supplied at their original stored resolution to API models.
GPT-5.4 uses the \emph{high} image-detail setting. For local Qwen inference,
images are converted to RGB and processed using the checkpoint's
\emph{AutoProcessor}. Qwen is evaluated with \emph{do\_sample=false},
\emph{bfloat16} precision, batch size 1, and automatic device placement.
The four subsets are distributed across four CUDA devices, with the subsets
evaluated independently.

\paragraph{Response Parsing.}
Yes/no responses are lowercased and mapped to \emph{yes} or \emph{no} when
the response begins with the corresponding word. If necessary, we accept the
answer when exactly one of the two words occurs in the response; ambiguous or
unparseable outputs are marked incorrect. Multiple-choice responses are parsed
as the first standalone letter from A to D.

For open-count questions, responses are lowercased, punctuation is removed,
and number words from zero to twenty are mapped to digits. We then extract the
first valid numeric count. Equivalent forms such as ``seven'' and ``7'' are
therefore treated identically, while explanations or other text do not receive
partial credit. Spatial attribute responses are normalized for capitalization,
spacing, and punctuation before exact matching. An output that cannot be
normalized to the required response format is counted as incorrect.

\paragraph{Scoring.}
Attribute, Language Elicitation, Counting, and Spatial are scored independently
at the sample level. Context and Texture contain four samples for each
Base--Edited case: two questions are asked about the Base image and two about
the Edited image. A case is correct only if all four responses are correct,
which we refer to as strict All4 accuracy. Language Elicitation is evaluated
using four-option multiple-choice accuracy, with the position of the
\emph{unknown} answer balanced across the dataset. Attribute and Counting use
normalized exact-count accuracy. Spatial uses normalized exact count or exact
attribute matching according to the question type. The overall
\name{}-Prior score is the unweighted macro-average of the Context, Texture,
Attribute, and Language Elicitation subset accuracies.

\paragraph{Confidence Intervals.}
We compute 95\% confidence intervals using a percentile bootstrap with 20,000
resamples and random seed 42. Resampling is performed over cases within each
subset. Thus, the four samples belonging to a Context or Texture case remain
grouped and are resampled together. For every bootstrap replicate, we
recompute the four subset accuracies and their unweighted macro-average. The
reported interval endpoints are the 2.5th and 97.5th percentiles of the
resulting bootstrap distributions.

\begin{table*}[t]
\centering
\small
\setlength{\tabcolsep}{4pt}
\renewcommand{\arraystretch}{1.08}
\begin{tabularx}{\textwidth}{p{2.8cm}p{3.2cm}p{3.0cm}>{\raggedright\arraybackslash}X}
\toprule
\textbf{Reported Name} &
\textbf{Model Identifier} &
\textbf{Interface} &
\textbf{Inference Configuration} \\
\midrule
Gemini 3.5 Flash &
\emph{gemini-3.5-flash} &
Google Gen AI API &
Temperature 0; thinking budget 0 \\

GPT-5.4 &
\emph{gpt-5.4} &
OpenAI Responses API &
Temperature 0; reasoning effort \emph{none}; image detail \emph{high} \\

Claude 4.6 &
\emph{claude-sonnet-4-6} &
OpenRouter &
Temperature 0 \\

Kimi-k2.6 &
\emph{kimi-k2.6} &
Official Moonshot API &
Thinking disabled; provider-supported deterministic configuration \\

Qwen 3.5 27B &
\emph{Qwen3.5-27B} &
Local Hugging Face Transformers &
Greedy decoding; thinking disabled; bfloat16 \\

Grok-4.3 &
\emph{grok-4.3} &
OpenRouter &
Temperature 0 \\
\bottomrule
\end{tabularx}
\caption{\textbf{Models and inference configurations used in our experiments.}
Greedy decoding is used whenever supported by the corresponding inference
interface.}
\label{tab:model_details}
\end{table*}

\section{Test Primers and Sample Specifications}
\label{app:test-primer}

A benchmark designer begins by providing a \emph{Test Primer}, which consists
of a natural-language Task Design written in Markdown and a Data Schema that
defines the required output fields and validation rules. The Task Design
describes the capability or failure mode to test, the visual construction
requirements, and the desired question format. The schema converts these
requirements into a machine-verifiable contract. A designer may use the
default schema provided by \name{} or replace it with a task-specific Pydantic
schema when additional fields or validation rules are required.

In our implementation, GPT-5.4 receives the Task Design together with the
schema and generates one structured Sample Specification at a time. The output
is accepted only if it satisfies the schema; invalid outputs and duplicate
concept identifiers are returned to the design model with validation feedback
for regeneration. Valid specifications are stored as JSONL records and passed
to the image-construction and question-building modules.

\paragraph{Default Data Schema.}
Table~\ref{tab:sample_spec_schema} summarizes the principal fields in the
default Sample Specification schema. The schema supports paired image editing
and single-image generation, as well as yes/no, multiple-choice, and
open-generation questions.

\begin{table*}[t]
\centering
\small
\setlength{\tabcolsep}{5pt}
\renewcommand{\arraystretch}{1.08}
\begin{tabularx}{\textwidth}{p{3.0cm}p{2.5cm}X}
\toprule
\textbf{Field} & \textbf{Type} & \textbf{Purpose} \\
\midrule
\emph{concept\_id}
& String
& Unique identifier used to reject duplicate specifications and connect
subsequent construction records. \\

\emph{pressure\_test\_type}
& String
& Name of the stress-test topic, such as \emph{context\_prior}. \\

\emph{objective}
& String
& Capability or failure mode that the specification is intended to test. \\

\emph{generation\_mode}
& Enum
& Either \emph{paired\_edit} or \emph{single\_image}. \\

\emph{base\_prompt}
& String
& Prompt used to construct the Base image for a paired intervention. \\

\emph{edit\_prompt}
& String
& Local transformation applied to the Base image to produce the Edited image. \\

\emph{image\_prompt}
& String
& Prompt used for tasks that require only one generated image. \\

\emph{task\_attributes}
& List
& Task-specific structured values, such as source entity, target entity,
material, count, or spatial relation. \\

\emph{questions}
& List
& Question specifications containing the probe identifier, image role,
question type, prompt, reference answer, evaluation rule, and optional
multiple-choice options. \\

\emph{review\_targets}
& List
& Instructions identifying the visual evidence and image region that a human
reviewer should verify. \\

\emph{generation\_notes}
& List
& Additional constraints used during image generation, editing, and review. \\
\bottomrule
\end{tabularx}
\caption{\textbf{Principal fields in the default Sample Specification schema.}}
\label{tab:sample_spec_schema}
\end{table*}
The schema also enforces cross-field constraints. A \emph{paired\_edit}
specification must contain both a Base prompt and an edit prompt, and its
questions may refer only to the Base or Edited image. A
\emph{single\_image} specification must instead contain an image prompt, and
all questions must refer to that image. Yes/no questions require
\emph{yes\_no\_exact} evaluation and a reference answer of either
\emph{yes} or \emph{no}; multiple-choice answers must identify one of the
provided options; and open-generation questions must specify an exact-match,
counting, or containment-based evaluation rule.

\paragraph{Context Task Design.}
We use the Context subset as a running example. Its Task Design instructs the
design model to construct a natural Base scene containing one contextually
expected source entity and no target entity. The Edited image must replace
only the source entity with a visually compatible but contextually unexpected
target while preserving the rest of the scene. The Task Design also defines
four complementary yes/no probes:

\begin{quote}
\small
\begin{enumerate}
    \item Source entity visible in Base: \emph{yes}.
    \item Target entity visible in Base: \emph{no}.
    \item Source entity visible in Edited: \emph{no}.
    \item Target entity visible in Edited: \emph{yes}.
\end{enumerate}
\end{quote}

The Markdown further requires both entities to remain human-recognizable while
occupying a minor region of a visually dense scene. It prohibits duplicate
source or target instances, readable labels, large isolated objects, obvious
editing artifacts, and changes outside the specified region.

\paragraph{Example Context Sample Specification.}
The following abbreviated record is an actual schema-compatible specification
generated for the Context task. Long image prompts are shortened only for
presentation \ref{fig:context_sample_spec}.

\begin{figure*}[t]
\centering
\begin{minipage}{0.96\textwidth}

\noindent
\textbf{Sample Specification}
\hfill
\texttt{Context / paired\_edit}
\vspace{2pt}

\begin{lstlisting}[style=specjson]
{
  "concept_id": "ctx01_repack_tape_bagel",
  "pressure_test_type": "context_prior",
  "generation_mode": "paired_edit",
  "objective": "Test whether a VLM follows visible evidence when an expected object is replaced by an unexpected object.",

  "base_prompt": "Generate a documentary photograph of a busy warehouse repacking floor. Include exactly one tan roll of masking tape on the lower shelf. No bagel or additional tape roll is visible. ...",

  "edit_prompt": "Replace only the masking tape with one plain bagel at the same location. Preserve all other objects, people, lighting, camera properties, and scene content. ...",

  "task_attributes": [
    {"name": "source_entity", "value": "roll of masking tape"},
    {"name": "target_entity", "value": "plain bagel"},
    {"name": "review_location", "value": "lower shelf of the central repacking bench"}
  ],

  "questions": [
    {"probe_id": "base_source", "image_role": "base", "question_type": "yes_no",
     "prompt": "Is a roll of masking tape visible?", "answer": "yes", "eval_type": "yes_no_exact"},

    {"probe_id": "base_target", "image_role": "base", "question_type": "yes_no",
     "prompt": "Is a plain bagel visible?", "answer": "no", "eval_type": "yes_no_exact"},

    {"probe_id": "edited_source", "image_role": "edited", "question_type": "yes_no",
     "prompt": "Is a roll of masking tape visible?", "answer": "no", "eval_type": "yes_no_exact"},

    {"probe_id": "edited_target", "image_role": "edited", "question_type": "yes_no",
     "prompt": "Is a plain bagel visible?", "answer": "yes", "eval_type": "yes_no_exact"}
  ],

  "review_targets": [
    {"label": "source object", "image_role": "base",
     "description": "Verify one tape roll at the specified location."},
    {"label": "target object", "image_role": "edited",
     "description": "Verify that one bagel replaces the tape roll."}
  ]
}
\end{lstlisting}

\end{minipage}

\caption{
\textbf{Abbreviated Context Sample Specification.}
The record connects visual construction, question definition, and human
verification within a schema-validated specification. Long prompts are
shortened for presentation.
}
\label{fig:context_sample_spec}
\end{figure*}
\paragraph{Question Instantiation.}
The questions are specified before image construction; the Question Builder
does not use another model to generate them. Once the required images are
available, it associates each question with the appropriate image through its
\emph{image\_role}, assigns a unique identifier, and exports the canonical
records used for filtering, human review, and final evaluation. 

For Context, the four probes share the same case identifier and Base--Edited
assets and are evaluated jointly using All4. Other subsets use the same
question-instantiation interface with their task-specific question formats and
scoring rules. Complete Test Primers and schemas will be released with the
benchmark-construction code.

\section{Task-Specific Instantiations}
\label{app:task-specific-instantiations}

We instantiate the pipeline on the four \name{}-Prior subsets and two
additional pilots. Each instantiation uses a different Test Primer, visual
construction requirement, question format, and validity criterion, while
retaining the same construction, filtering, verification, repair, and export
workflow.

Table~\ref{tab:task_specific_instantiations} summarizes these settings. Context
and Texture use paired Base--Edited cases, whereas Attribute and Language
Elicitation use a single final image. Counting and Spatial are evaluated at the
individual-image level and do not use the All4 case-level metric.

\begin{table*}[t]
\centering
\small
\setlength{\tabcolsep}{3pt}
\renewcommand{\arraystretch}{1.05}
\begin{tabularx}{\textwidth}{p{2.1cm}p{1.9cm}p{3.5cm}p{2.2cm}>{\raggedright\arraybackslash}X}
\toprule
\textbf{Instantiation}
& \textbf{Image Mode}
& \textbf{Task-Specific Construction}
& \textbf{Question Format}
& \textbf{Primary Validity Requirement} \\
\midrule

Context
& Paired Base--Edited
& Replace one contextually expected source entity with a visually compatible
but contextually unexpected target.
& Four yes/no probes
& The source appears only in Base, the target appears only in Edited, and all
non-target content is preserved. \\

Texture
& Paired Base--Edited
& Change the visible surface of a familiar entity from its canonical material
to a counterfactual material while preserving its identity and geometry.
& Four yes/no probes
& Canonical and counterfactual surface cues are visually distinguishable, and
the intervention does not change the entity or surrounding scene. \\

Attribute
& Single final image
& Construct a familiar entity with a noncanonical number of visible components
while preserving its category identity.
& Open-ended count
& Every counted component is visible and individually distinguishable, and
the annotated count matches the image exactly. \\

Language Elicitation
& Single image
& Construct a scene in which the question suggests a plausible answer although
the queried evidence is hidden, occluded, unreadable, or outside the frame.
& Four-option MCQ
& The requested information cannot be determined from the image, making
\emph{unknown} the uniquely supported answer. \\

Counting
& Independently scored images
& Construct dense scenes containing many instances of a precisely defined
target category together with visually similar distractors.
& Open-ended count
& Every target instance has a visible boundary, distractors are
distinguishable, and the reference count is exact. \\

Spatial
& Independently scored images
& Construct dense multi-layer 3D lattices with controlled object colors,
shapes, coordinates, and spatial relations.
& Open-ended attribute or count
& Each question has one visually verifiable answer under an explicitly defined
coordinate system or spatial constraint. \\

\bottomrule
\end{tabularx}
\caption{\textbf{Task-specific instantiations of the shared \name{} pipeline.}
The main \name{}-Prior benchmark contains Context, Texture, Attribute, and
Language Elicitation. Counting and Spatial are reported separately as
additional pipeline instantiations.}
\label{tab:task_specific_instantiations}
\end{table*}
\paragraph{Context.}
A Context specification defines a dense, coherent Base scene containing
exactly one source entity that is natural for the scene and no instance of the
target entity. The source is placed in a precisely described review location
and remains identifiable without becoming the focal object. The target is
selected to have a similar apparent size, orientation, color, or silhouette so
that it can occupy the same location after editing while remaining
contextually unexpected.

The edit instruction replaces only the source with the target and preserves
the camera, people, lighting, composition, surrounding objects, and local
occlusion pattern. The specification contains four question records:
\emph{base\_source}, \emph{base\_target},
\emph{edited\_source}, and \emph{edited\_target}, with reference answers
\emph{yes}, \emph{no}, \emph{no}, and \emph{yes}, respectively. A candidate is
visually valid only when both entities are human-recognizable in their intended
images, neither entity appears in the wrong image, and no unrelated scene
content changes in a way that affects the questions.

\paragraph{Texture.}
A Texture specification begins with a familiar entity displaying its expected
surface, such as a plastic watering can or a wooden rolling pin. The
counterfactual material is chosen to conflict with the entity's canonical
appearance while providing recognizable visual evidence, such as visible
fabric fibers, foam structure, glass transparency, or a paper-like surface.
The intervention changes the visible material of the target entity but
preserves its category, geometry, scale, pose, and location.

The four probes ask whether the entity has the canonical or counterfactual
surface in the Base and Edited images. Their answer pattern is again
\emph{yes/no} for Base and \emph{no/yes} for Edited. Human validation checks
that the surface change covers the intended visible region, that the material
cues are sufficiently clear for direct visual judgment, and that the edit
does not deform the entity or alter neighboring objects.

\paragraph{Attribute.}
Attribute specifications target familiar objects whose components normally
follow a canonical count, including animal legs, chair legs, shirt buttons,
instrument holes, and repeated symbols on playing cards. The image is
constructed or edited to contain a noncanonical number of visible components
while retaining the object's recognizable identity. The intervention must
produce complete components rather than duplicated fragments, merged
structures, shadows, or ambiguous partial instances.

Each exported image is associated with one open-ended question asking for the
visible component count. The reference answer is derived from the verified
image rather than assumed from the construction prompt. During validation,
the reviewer confirms the object identity, checks each visible component, and
ensures that occlusion or malformed geometry does not make the count
ambiguous.

\paragraph{Language Elicitation.}
Language Elicitation specifications construct scenes in which contextual
wording makes one answer appear plausible although the necessary evidence is
not available in the image. Examples include asking about the contents of a
closed opaque container, text on a face-down object, information hidden by
glare or blur, or content outside the image boundary. The image prompt must
ensure that the queried information is genuinely unavailable and that no
incidental visual cue reveals the intended distractor.

Each question contains three plausible content answers and one
\emph{unknown} option. The plausible alternatives are selected to reflect
answers that may be suggested by the scene or question wording, rather than
visually supported answers. The position of \emph{unknown} is shuffled and
balanced across the subset. A candidate is valid only when none of the three
content options can be confirmed from the image and \emph{unknown} is
therefore the uniquely supported response.

\paragraph{Counting.}
Counting specifications define a target category using visible properties
such as object type, color, pattern, or location. The scene contains dozens of
target instances distributed across natural clusters, together with
distractors sharing selected properties with the target. For example, a scene
may contain yellow lemons among other yellow or round objects, or
four-hole blue buttons among buttons and beads with similar colors and shapes.
Targets are not arranged in regular rows and may touch or overlap slightly,
but each instance must retain a visible boundary.

The construction process may produce Base and Edited images with different
target counts. The edit adds or removes a specified number of targets while
preserving the scene and distractors. Unlike Context and Texture, each image
is exported as an independent open-count sample. Human verification determines
the final visible count and rejects images containing merged, duplicated,
heavily occluded, or otherwise uncountable target instances.

\paragraph{Spatial.}
Spatial specifications construct dense multi-layer 3D lattices containing
repeated colored shapes at controlled cells. A fixed coordinate convention
defines the front-left-bottom cell as the origin cell, with the horizontal
axis increasing from left to right, depth increasing from front to back, and
height increasing from bottom to top. The same convention is stated explicitly
in every coordinate question to avoid dependence on an unstated viewpoint.

The task supports coordinate lookup questions, such as identifying the color
and shape at a specified cell, as well as constrained counting questions, such
as counting red cubes on the top layer. Specifications may also move a target
object between cells while preserving the lattice, camera, and all other
objects. As with Counting, Base and Edited images are evaluated independently.
A candidate is valid only when the lattice orientation is visually
recoverable, the queried cells or layers remain visible, the target objects
are distinguishable from distractors, and the reference answer can be
verified without relying on the generation prompt.

\begin{figure*}[t!]
    \centering
    \includegraphics[width=0.95\textwidth]{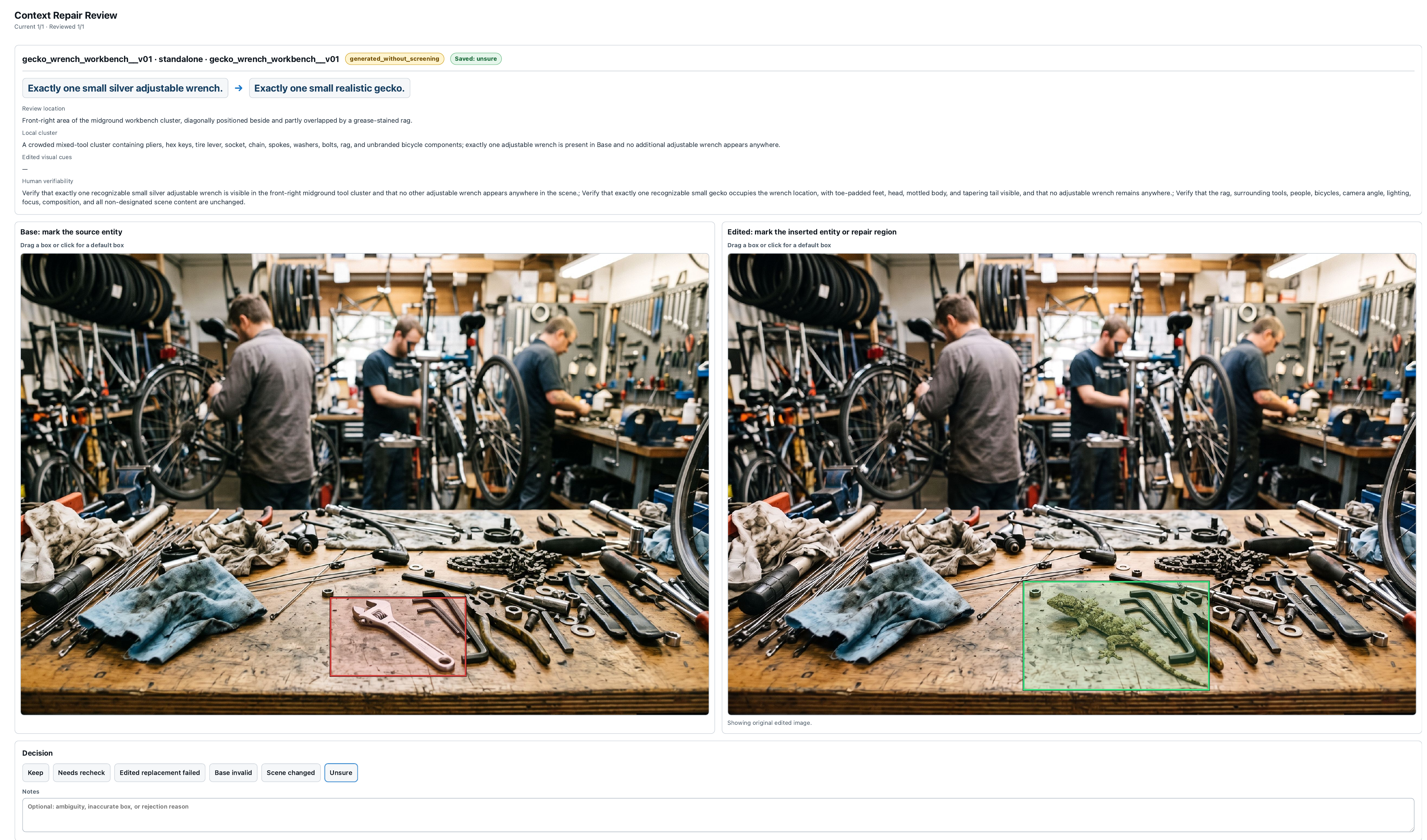}
    \caption{Overview of the \name{} annonation platform.}
    \label{fig:review_platform}
\end{figure*}

\section{Human Verification and Curation Platform}
\label{app:human-verification}

\subsection{Annotation Platform}

Figure~\ref{fig:review_platform} shows the platform used to verify candidates
retained by automated pressure filtering. For paired-image cases, the Base and
Edited images are displayed side by side together with the intended source-to-
target transformation and a natural-language description of the review
location. This location cue helps reviewers find the relevant evidence in
visually dense scenes. Reviewers mark the source entity in the Base image and
the inserted entity or edited region in the Edited image using bounding boxes.
The normalized box coordinates are stored as curation metadata, allowing the
regions to be highlighted during later inspection and reused for localized
repair. These annotations are not provided to the evaluated VLMs. Reviewers inspect whether the Base image is valid, whether the intended
transformation is correctly realized, and whether unrelated scene content is
preserved. The interface also presents the four pressure-screening probes,
including their reference answers, filtering-model predictions, and
correctness. A valid case is marked as \emph{Keep}; otherwise, it can be marked
as requiring reinspection, having a failed edit, having an invalid Base image,
containing unintended scene changes, or being uncertain. Reviewers may also
record notes explaining ambiguous cases or rejection decisions. The platform additionally provides localized image repair and real-image
authoring tools. Reviewers can select a defective region in an Edited image,
generate a repaired candidate, compare it with the current image, and either
accept the repair or restore the original version. The real-image authoring
interface supports importing photographs, specifying source and target
entities, constructing edited candidates, defining question--answer pairs, and
exporting the resulting annotations in the same structured format.

\subsection{Localized Image Repair}
\label{app:localized-repair}

\begin{figure*}[t]
    \centering
    \includegraphics[width=\textwidth]{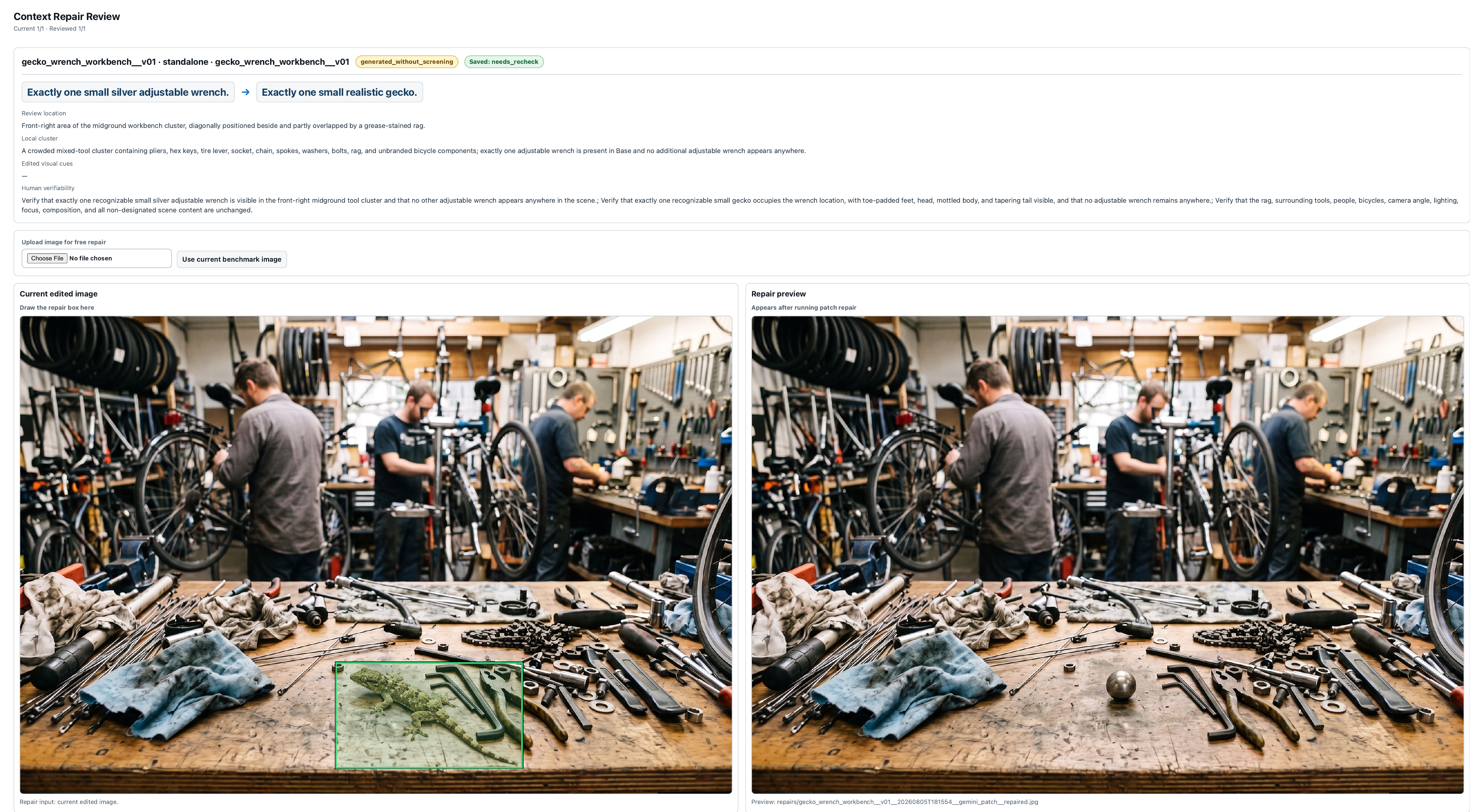}
    \vspace{0.5em}

    \includegraphics[width=\textwidth]{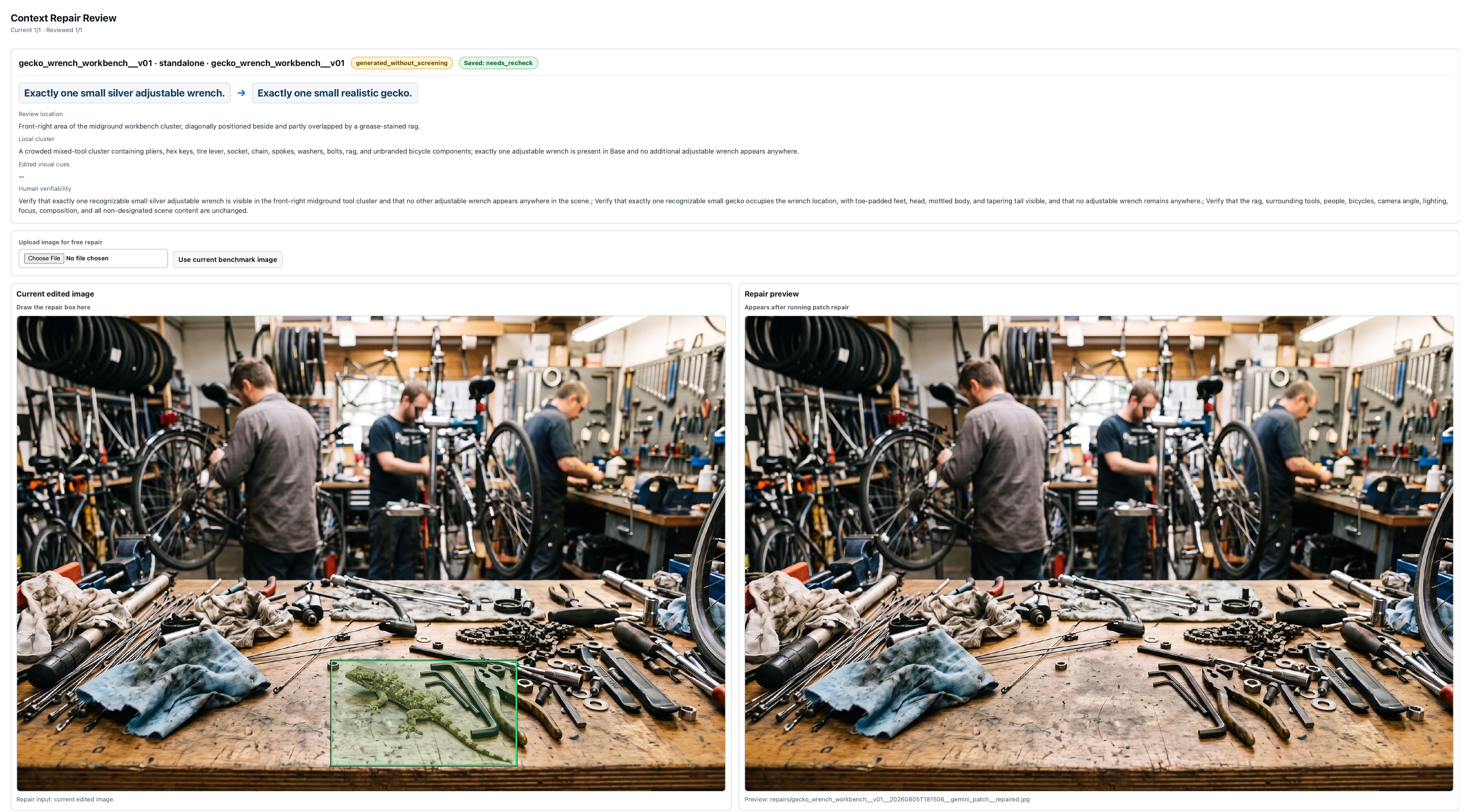}
    \caption{
    \textbf{Localized image repair in the curation platform.}
    Top: \emph{Swap Entity} replaces the bagel inside the selected region with
    a ball. Bottom: \emph{Remove Entity} removes the bagel and reconstructs the
    surrounding tabletop. The current Edited image is shown on the left and
    the repair preview on the right. Reviewers can accept the repaired image or
    revert to the original Edited image.
    }
    \label{fig:localized_repair}
\end{figure*}

Fine-grained full-image editing is frequently unreliable, especially when the
requested change affects only a small object in a visually complex scene. A
generation model may ignore the requested change, leave visible remnants of
the original entity, or modify unrelated objects and scene content. Repeatedly
regenerating the complete image can therefore destroy otherwise valid visual
evidence and weaken the controlled relationship between the Base and Edited
images. To address this problem, our curation platform provides localized patch repair,
as illustrated in Figure~\ref{fig:localized_repair}. When a reviewer identifies
a defective region, they draw a bounding box around the failed edit or local
artifact and select a repair operation. The system expands the selected region
slightly to retain its surrounding visual context, crops the resulting patch,
and sends only this marked patch to the image generation model together with a
task-specific repair instruction. The repaired patch is resized when necessary
and blended back into the original image using a soft mask. This procedure
preserves the image outside the local neighborhood and reduces unintended
changes to unrelated scene content.

The platform supports four repair operations: removing an unwanted entity,
swapping one entity for another, cleaning local visual artifacts, and restoring
the background. Figure~\ref{fig:localized_repair} presents the two entity-level
operations. For \emph{Swap Entity}, the reviewer specifies both the source and
target entities; for \emph{Remove Entity}, the selected object is removed and
the exposed region is completed using its local context. Reviewers may further
refine the repair instruction to specify expected appearance, scale, lighting,
or content preservation constraints.

A generated repair is treated as a candidate rather than being accepted
automatically. The platform displays the original Edited image and repair
preview side by side, marks the case as requiring reinspection, and allows the
reviewer to accept the repair or return to the previous image. When a repair is
accepted, the previous Edited image is retained as a backup and the repair
operation is recorded. The repaired case must then pass human verification
before it is included in the benchmark.

\subsection{Blind User Study of Localized Repair Quality}
\label{app:repair-user-study}

\begin{figure*}[t!]
    \centering
    \includegraphics[width=0.95\textwidth]{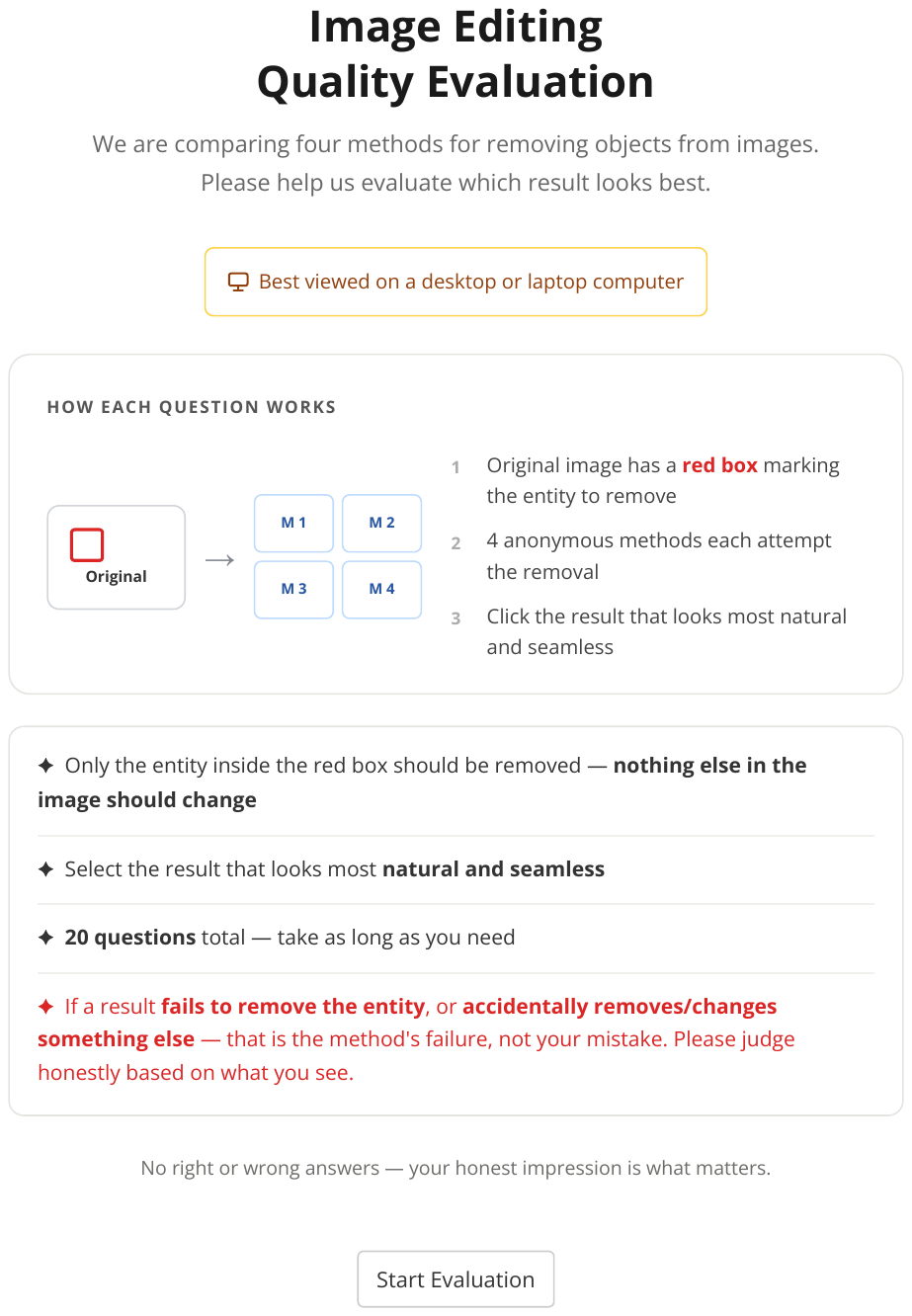}
    \caption{
    \textbf{Instructions for the method-anonymous repair-quality study.}
    Participants are asked to select the most natural result that removes only
    the marked entity while preserving all unrelated image content.
    }
    \label{fig:repair_questionnaire_instructions}
\end{figure*}

\begin{table*}[!t]
\centering
\small
\setlength{\tabcolsep}{3pt}
\renewcommand{\arraystretch}{1.08}
\begin{tabular*}{\textwidth}{@{\extracolsep{\fill}}llrrrrrr}
\toprule
\textbf{Subset} & \textbf{Metric}
& \textbf{Gemini 3.5 Flash}
& \textbf{GPT-5.4}
& \textbf{Claude 4.6}
& \textbf{Kimi-k2.6}
& \textbf{Qwen 3.5 27B}
& \textbf{Grok-4.3} \\
\midrule

\multirow{6}{*}{Context Prior}
& \textbf{All4 (P)} $\uparrow$ & 0  & 1  & \textbf{10} & 7  & 3  & 4  \\
& Edit2 $\uparrow$             & 0  & 2  & \textbf{13} & 9  & 7  & 6  \\
& Q1 Base+ $\uparrow$          & 97 & \textbf{99} & 72 & 85 & 93 & 97 \\
& Q2 Base- $\uparrow$          & 82 & 67 & \textbf{94} & 84 & 85 & 62 \\
& Q3 Edit- $\uparrow$          & 0  & 3  & \textbf{33} & 17 & 10 & 7  \\
& Q4 Edit+ $\uparrow$          & 86 & 82 & 52 & 58 & 68 & \textbf{88} \\
\midrule

\multirow{6}{*}{Texture}
& \textbf{All4 (P)} $\uparrow$ & \textbf{52} & 28 & 40 & \textbf{52} & 46 & 28 \\
& Edit2 $\uparrow$             & 52 & 28 & 41 & \textbf{54} & 47 & 29 \\
& Q1 Base+ $\uparrow$          & 99 & \textbf{100} & 98 & 98 & 99 & 99 \\
& Q2 Base- $\uparrow$          & \textbf{100} & 99 & \textbf{100} & 99 & \textbf{100} & \textbf{100} \\
& Q3 Edit- $\uparrow$          & \textbf{81} & 30 & 54 & 69 & 63 & 38 \\
& Q4 Edit+ $\uparrow$          & 58 & \textbf{82} & 65 & 67 & 65 & 65 \\
\midrule

Attribute
& \textbf{Count Acc. (P)} $\uparrow$
& \textbf{26} & 20 & 17 & 17 & 14 & 16 \\
\midrule

Language Elicitation
& \textbf{MCQ Acc. (P)} $\uparrow$
& 11 & 23 & \textbf{58} & 17 & 29 & 23 \\
\midrule

Overall
& \textbf{Macro Avg.} $\uparrow$
& 22.3 & 18.0 & \textbf{31.3} & 23.3 & 23.0 & 17.8 \\

\bottomrule
\end{tabular*}
\caption{
\textbf{Detailed results on \name{}-Prior.}
Rows marked with (P) are the primary subset metrics used to compute the
macro-average. For Context Prior and Texture, All4 requires all four
Base--Edited probes to be correct, while Edit2 requires both Edited-image
probes to be correct. Q1 and Q2 respectively test the expected and
counterfactual evidence in the Base image; Q3 and Q4 test the same evidence
after editing, with expected answers \emph{yes/no/no/yes}. Attribute reports
normalized exact-match counting accuracy, and Language Elicitation reports
four-option multiple-choice accuracy. All values are percentages. Bold values
indicate the highest score in each row.
}
\label{tab:detailed_results}
\vspace{-0.5em}
\end{table*}
We conduct a randomized, method-anonymous user study to evaluate the perceptual
quality of images produced by the localized repair tool. The survey interface are shown in
Figures~\ref{fig:repair_questionnaire_instructions}. The study contains 20 image-repair
cases. Each question first presents an original image with a red bounding box
indicating the entity to be removed, followed by four candidate repair results.
Participants select the result that most naturally and seamlessly realizes the
requested edit. The instructions emphasize that the marked entity should be
completely removed while unrelated image content should remain unchanged.

We compare four methods: Gemini prompt-only full-image editing, OpenCV
inpainting, Gemini patch generation with hard pasting, and our Gemini patch
generation with soft-mask blending. The hard-paste baseline and our method use
the same generated patch, isolating the effect of the compositing strategy.
Candidate results are displayed only as \emph{Method 1}--\emph{Method 4},
without revealing their underlying methods. Their display positions are
independently shuffled for every question, and the system records both the
randomized order and the selected method. Each question requires one selection
and cannot be skipped.

Forty survey sessions were initiated. We analyze the 20 sessions that completed
all 20 required comparisons, yielding 400 valid selections. We use entity
removal throughout the study to maintain a consistent and unambiguous
evaluation objective. The study therefore directly evaluates the removal
setting and the shared patch-generation and compositing components of the
localized repair mechanism. Entity swapping is demonstrated qualitatively in
Figure~\ref{fig:localized_repair} but is not included in this user study.

\begin{table}[t]
\centering
\footnotesize
\setlength{\tabcolsep}{4pt}
\renewcommand{\arraystretch}{1.08}
\begin{tabular}{lrr}
\toprule
\textbf{Repair Method} & \textbf{Votes} & \textbf{Preference (\%)} \\
\midrule
Gemini full-image prompt  & 16  & 4.0 \\
OpenCV inpainting         & 0   & 0.0 \\
Gemini patch + hard paste & 10  & 2.5 \\
\textbf{Ours: patch + soft blend}
                          & \textbf{374} & \textbf{93.5} \\
\bottomrule
\end{tabular}
\caption{
Method preferences from 20 participants over 20 image-repair cases, yielding
400 selections in total.
}
\label{tab:repair_user_study}
\end{table}

As shown in Table~\ref{tab:repair_user_study}, our method is selected in 374 of
the 400 comparisons, corresponding to an overall preference rate of 93.5\%.
Because every participant completes the same 20 questions, this is also the
mean participant-level preference rate. The standard deviation across
participants is 8.1 percentage points, with individual preference rates ranging
from 70\% to 100\%. Participants therefore show a clear preference for
localized patch generation with soft-mask blending over full-image editing,
OpenCV inpainting, and hard patch compositing on the tested entity-removal
cases.

\subsection{Illustrative Pipeline Cost}
\label{app:pipeline-cost}

Exact historical costs cannot be recovered because early task exploration and
benchmark construction shared the same API accounts. We therefore report an
illustrative marginal-cost estimate for the stabilized Context recipe using
the Gemini-based image pipeline. For every 100 initial candidates, we estimate
that approximately 50 pass pressure filtering and enter human review, of which
30 are accepted directly, 10 are accepted after localized repair, and 10 are
rejected. This yields approximately 40 retained cases.

\begin{table*}[t]
\centering
\small
\setlength{\tabcolsep}{4pt}
\renewcommand{\arraystretch}{1.05}
\begin{tabularx}{\textwidth}{p{2.3cm}p{2.2cm}X p{2.0cm}p{2.4cm}}
\toprule
\textbf{Stage} &
\textbf{Model} &
\textbf{100 Initial Candidates} &
\textbf{Estimated Cost} &
\textbf{Cost per 100 Retained} \\
\midrule
Sample specification
& GPT-5.4
& 100 structured outputs
& \$2.0--\$5.0
& \$5.0--\$12.5 \\

Base-image generation
& Gemini 3.1 Flash Image
& 100 image generations
& \$6.7
& \$16.8 \\

Targeted image editing
& Gemini 3.1 Flash Image
& 100 image edits
& \$6.7
& \$16.8 \\

Pressure filtering
& Gemini 3.5 Flash
& 400 image--question evaluations
& \$1.1--\$2.2
& \$2.8--\$5.5 \\

Localized repair
& Gemini 3.1 Flash Image
& approximately 10 repair calls
& \$0.7--\$1.4
& \$1.8--\$3.5 \\
\midrule
\textbf{Total API cost}
& --
& --
& \textbf{\$17.2--\$22.0}
& \textbf{\$43.0--\$55.0} \\
\midrule
Human verification
& --
& 50 reviewed cases
& 25--50 min
& 1.0--2.1 h \\

Repair and reinspection
& --
& approximately 10 cases
& 10--20 min
& 0.4--0.8 h \\
\midrule
\textbf{Total active human time}
& --
& --
& \textbf{35--70 min}
& \textbf{1.5--2.9 h} \\
\bottomrule
\end{tabularx}
\caption{
\textbf{Illustrative marginal cost of the stabilized Context recipe.}
Image costs assume 1K outputs from Gemini 3.1 Flash Image. The estimate includes
sample-specification generation, Base-image generation, targeted editing,
pressure filtering, localized repair, and active human review. It excludes
one-time task exploration, prompt development, framework engineering, and final
benchmark evaluation. Costs and retention rates vary with the task
specification and required visual intervention.
}
\label{tab:pipeline_cost}
\end{table*}
\section{Additional Results}

Table~\ref{tab:detailed_results} localizes the failures within the paired-image
subsets. Base-image accuracy is generally high, whereas performance drops after
editing. Indicating that models may recognize the inserted target while
still reporting the replaced source. These results motivate the strict
All4 metric over individual-probe accuracy.

\end{document}